\documentclass[11pt, a4paper, logo, copyright, nonumbering]{xiaomi}

\usepackage[numbers, sort&compress, square]{natbib}
\usepackage{dblfloatfix}
\usepackage[normalem]{ulem}
\usepackage{caption}
\usepackage{dramatist}
\usepackage{xspace}
\usepackage{pifont} 
\usepackage{multirow}
\usepackage{tcolorbox}
\usepackage{xltabular}
\usepackage{longtable}
\usepackage{hyperref}
\usepackage{wrapfig}

\usepackage{amsfonts}
\usepackage{amsmath}
\usepackage{amssymb}
\usepackage{lineno}
\usepackage{multirow}
\usepackage{adjustbox}

\usepackage[bottom]{footmisc}

\usepackage{CJKutf8}
\usepackage{setspace}
\usepackage{makecell}
\usepackage{graphicx}
\usepackage{subcaption}
\usepackage{multicol} 

\usepackage[utf8]{inputenc} 
\usepackage[T1]{fontenc}    
\usepackage{hyperref}
\usepackage{cleveref}
\usepackage{url}            
\usepackage{booktabs}       
\usepackage{amsfonts}       
\usepackage{nicefrac}       
\usepackage{microtype}      
\usepackage{xcolor}         
\usepackage{colortbl}
\usepackage{tcolorbox}
\usepackage{xspace}
\definecolor{BrickRed}{rgb}{.72,0,0}
\definecolor{darkgreen}{rgb}{0.0, 0.5, 0.0}
\definecolor{ForestGreen}{RGB}{34,139,34}
\definecolor{LakeBlue}{RGB}{0,61,153}
\definecolor{MiOrange}{RGB}{255,225,204}
\definecolor{Hex}{RGB}{225,213,231}

\hypersetup{
    colorlinks=true,
    allcolors=LakeBlue
}

\usepackage{siunitx}
\usepackage[table]{xcolor}
\definecolor{best}{RGB}{255,0,0}      
\definecolor{second}{RGB}{0,0,255}    

\title{\centering TIGER: Taming Identity, Geometry, and Generative Priors for High-Quality Face Video Restoration}

\titlerunning{High-Quality Video Face Restoration}

\author{
  Yang Zhou$^1$$^*$,
  Wenxue Li$^{1,2}$$^*$,
  Peng Zhang$^1$$^\dagger$,
  Yifei Chen$^1$,
  Fei Wang$^1$,
  Daiguo Zhou$^1$
}
\institute{
$^1$ MiLM Plus, Xiaomi Inc.\\
$^2$ The Hong Kong University of Science and Technology (Guangzhou) \\
}
\footnotetext[1]{$^*$Equal contribution.}
\footnotetext[2]{$^\dagger$Corresponding author.}

\begin{document}

\begin{abstract}
Face Video Restoration (FVR) aims to recover high-fidelity facial videos from degraded input while preserving identity and semantic consistency across frames. Existing methods often struggle to simultaneously address three key challenges: identity shift, viewpoint-entangled guidance, and perceptual realism.
To tackle these issues, we propose TIGER, a structured tri-prior fusion framework that \textbf{T}ames \textbf{I}dentity, \textbf{G}eometry, and g\textbf{E}nerative p\textbf{R}iors for high-quality FVR. 
Specifically, an \textit{Identity Prior} is first established by injecting subject-discriminative embeddings into the latent space, effectively anchoring the subject’s identity against severe degradations. 
Then, to provide temporally consistent structural guidance for dynamic videos, TIGER constructs a \textit{Geometry Prior} by lifting 2D reference cues into a disentangled 3D parameter space, creating a geometric anchor through cross-source parameter fusion. 
Moreover, to achieve maximum efficiency without compromising realism, we harness the video generation model's \textit{Generative Prior} through a one-step rectified flow. 
We further design a progressive three-stage training optimazition strategy that refines structural fidelity, textural reconstruction, and distribution-level realism to ensure robust optimization.
We also construct a large-scale FVR dataset to facilitate robust training and standardized evaluation.
Extensive experiments demonstrate that TIGER achieves state-of-the-art performance in both identity fidelity and temporal stability, delivering a high-quality, efficient and identity-consistent FVR.
Project page: \url{https://yzhoulv.github.io/Tiger/}.
\end{abstract}

\maketitle

\begin{figure*}[t!]
    \centering
    \includegraphics[width=1.0\linewidth]{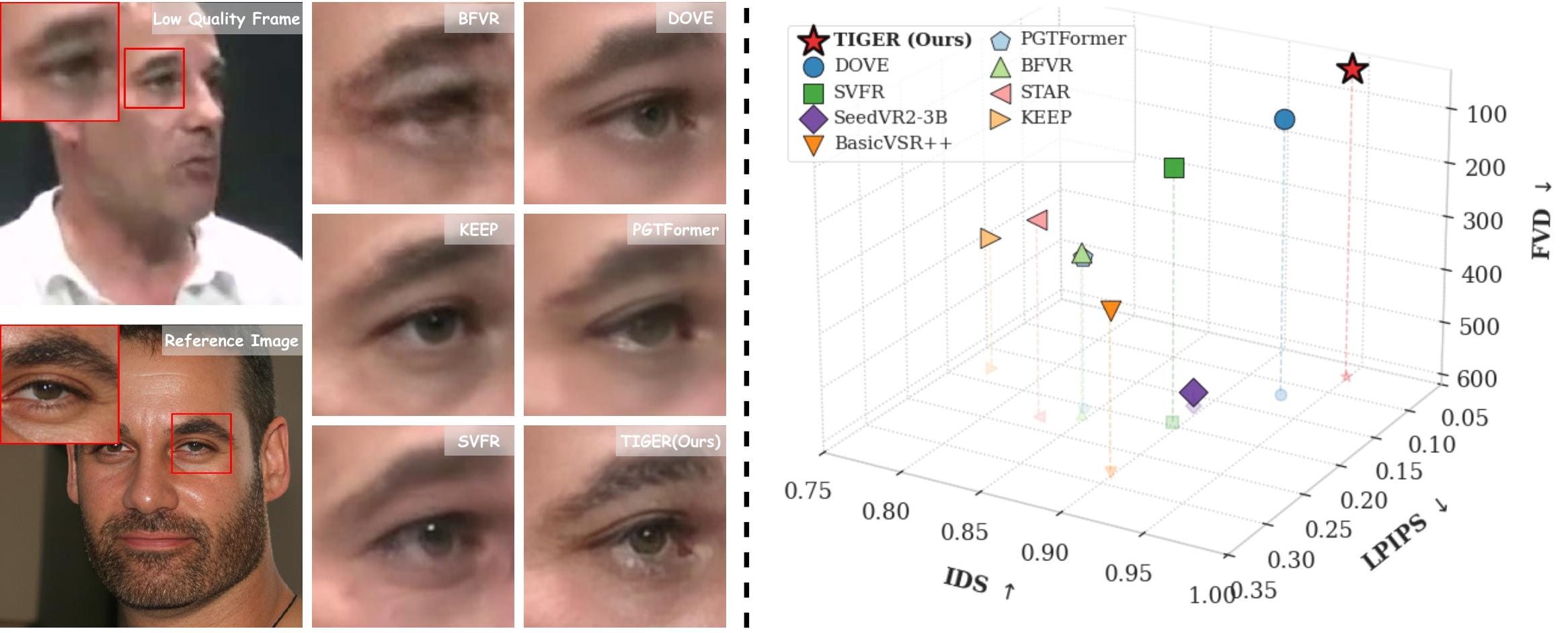}
    \caption{Qualitative and quantitative comparisons with state-of-the-art video restoration methods. \textbf{Left}: Visual comparison of restored facial details, particularly eye regions, demonstrating that TIGER generates more realistic and perceptually pleasing results. \textbf{Right}: 3D scatter plot showing the trade-offs among IDS (identity similarity), LPIPS (perceptual quality), and FVD (temporal consistency). TIGER achieves superior performance in both visual quality and identity preservation.}
    \label{fig:abstract}
\end{figure*}

\section{Introduction}
\label{sec:intro}

Recent advances in video restoration have demonstrated remarkable progress in reconstructing high-fidelity content from severely degraded inputs~\cite{Xie_2025_STAR,chan2022basicvsr++,chan2022realbasicvsr}. 
Among these tasks, Face Video Restoration (FVR) is particularly challenging due to its structural and semantic constraints.
Unlike generic video restoration, FVR must simultaneously recover fine-grained facial details, preserve the subject’s identity-defining characteristics, and maintain temporal coherence across dynamic expressions and pose variations~\cite{han2025show,zhou2022codeformer}. 
The difficulty is further amplified when degradations are severe, as the available observations become insufficient to reliably preserve identity.

The advent of powerful generative foundation models \cite{rombach2022ldm,wan2025wan} has shifted the paradigm from mere degradation removal to photorealistic synthesis, significantly boosting perceptual quality \cite{he2024venhancer,yeh2024diffir2vr}. 
However, naively adapting these generative paradigms to FVR reveals three critical failures that constitute a formidable trilemma:
\textit{\textbf{(1) Identity Drift:}} 
Under severe degradations, identity-discriminative cues are heavily corrupted.
Consequently, strong generative priors tend to dominate the reconstruction process, producing visually plausible yet subject-inconsistent facial traits.
\textit{\textbf{(2) Viewpoint-Entangled Guidance:}} While reference images can anchor identity, a static 2D reference inherently entangles identity with a single pose and expression.
When large viewpoint or motion variations occur, the model must extrapolate unseen structures from learned appearance statistics, reintroducing identity ambiguity and temporal inconsistency.
\textit{\textbf{(3) Perceptual Quality and Realism:}} 
To achieve efficient inference, many one-step acceleration techniques rely on distillation or regression-based objectives, which suffer from a regression-to-the-mean effect, resulting in softened textures and diminished photorealistic realism. Balancing generative realism with computational efficiency therefore remains an open challenge.

To resolve this trilemma, we propose \textbf{TIGER}, a structured framework that \textbf{T}ames \textbf{I}dentity, \textbf{G}eometry, and g\textbf{E}nerative p\textbf{R}iors for high-quality FVR.
We reformulate FVR as prior-conditioned latent transport, where identity cues, geometric structure, and generative knowledge are utilized to stabilize identity, promote geometry-consistent temporal dynamics, and preserve perceptual richness within a unified framework. 

Specifically, we first establish a \textbf{\textit{Identity Prior}} by leveraging ArcFace-based \cite{deng2019arcface} identity embeddings extracted from the reference image. By injecting the subject-discriminative anchors into the diffusion backbone, TIGER ensures that the synthesized features remain unequivocally bound to the target subject's unique traits, effectively eliminating identity drift.
To bridge the viewpoint-entangled guidance gap, we lift 2D reference cues into a 3D parameter space to construct a \textbf{\textit{Geometry Prior}}. By performing cross-source parameter fusion—recombining the static identity of the reference with the frame-specific motion of the degraded sequence—we generate a sequence of viewpoint-agnostic normal maps. This geometry prior serves as a dense, temporally consistent structural scaffold, ensuring that the latent transport process is strictly bounded by 3D-aware facial dynamics. 
Finally, we harness the foundation model's \textbf{\textit{Generative Prior}} by formulating the restoration as a one-step rectified flow. Guided by the aforementioned geometry and identity priors, TIGER learns a direct velocity field that transports the degraded latent to the high-fidelity manifold in a single inference step. 
To ensure the robust optimization of this synergistic prior integration, we design a progressive three-stage training strategy that sequentially refines structural fidelity, textural reconstruction, and holistic realism.

Our main contributions are:
\begin{itemize}
\item We propose TIGER, a FVR framework that effectively resolves the long-standing trilemma between identity fidelity, viewpoint-entangled guidance, and perceptual realism through the integration of identity, geometry, and generative priors.

\item We introduce a prior-conditioned one-step rectified-flow formulation. Equipped with cross-source 3D parameter fusion and hybrid prior injection, it achieves high-fidelity, identity-preserving restoration in a single forward pass.

\item We design a progressive three-stage optimization strategy that sequentially enhances structural fidelity, fine-grained detail, and distribution-level realism.

\item To facilitate robust training and standardized evaluation, we construct a new large-scale FVR dataset comprising 28,491 training samples and establish a comprehensive benchmark. Extensive experiments demonstrate that TIGER achieves state-of-the-art performance. 

\end{itemize}

\section{Related Work}

\subsection{General Video Enhancement}  
Early video super-resolution (VSR) and video restoration methods primarily relied on convolutional networks \cite{chan2021basicvsr,edvr}, optical flow–based alignment \cite{guo2024generalizable}, and later, Transformer-based designs \cite{liang2022vrt}. To handle real-world degradations, subsequent approaches incorporated pre-cleaning modules \cite{chan2022realbasicvsr}, degradation kernel estimation \cite{ji2020real}, synthetic degradation pipelines \cite{chang2020learning,song2024negvsr}, and GAN-based objectives \cite{chen2024high}. Temporal consistency was further enforced through alignment error minimization or motion compensation \cite{lei2020blind}.

More recently, diffusion models—leveraging strong generative priors—have been adopted for image and video restoration \cite{li2022srdiff,saharia2022image,lugmayr2022repaint,weng2024vires}. In VSR, this paradigm has led to methods employing diffusion backbones \cite{rota2024stablevsr,li2025diffvsr}, high-capacity GANs \cite{xu2024videogigagan}, autoregressive generation \cite{sun2025ardiffusion,zhang2025infvsr}, and explicit temporal modeling \cite{wang2025temporal}. To address the computational burden of iterative sampling, acceleration techniques such as distillation \cite{meng2023distillation,zhou2024simple} and efficient stochastic differential equation (SDE) solvers \cite{lu2025dpm,zheng2023dpmsolver} have been proposed. Despite these advances, general VSR frameworks often lack  identity-aware modeling, resulting in identity drift and suboptimal performance in portrait-centric scenarios.

\subsection{Face Video Enhancement}

Blind FVR requires simultaneously preserving identity, recovering fine textures, and maintaining temporal stability under unknown and severe degradations. 
However, general VSR models lack explicit facial priors and often exhibit identity inconsistency, texture artifacts, and temporal flickering. 
This limitation has motivated a shift toward face-specific designs. 
Early efforts enhance inter-frame correlation through temporal modeling \cite{feng2024keep}, while more recent approaches integrate generative priors with facial semantics. For example, some methods learn identity-aware representations end-to-end without explicit alignment \cite{pgtformer}; others incorporate motion or diffusion priors to stabilize dynamic details \cite{wang2025svfr}, or leverage structured codebooks to improve the robustness of feature representations \cite{xie2025bfvr}.
Moreover, model performance remains bottlenecked by training data quality. Widely used datasets like VoxCeleb \cite{Voxceleb} and CelebV-HQ \cite{zhu2022celebvhq}, though large, contain in-the-wild degradations (e.g., motion blur, occlusion, and low illumination) that hinder the learning of clean facial distributions. 
This degraded-to-clean learning gap introduces a distribution mismatch that limits the recovery of realistic high-frequency details, highlighting the need for higher-fidelity training data.

\subsection{Reference-Guided Face Restoration}  
Incorporating identity information from a reference image has emerged as a promising strategy to enhance identity fidelity in blind restoration. Early methods transferred identity cues via pose alignment or keypoint matching \cite{li2018gfrnet,li2020asfnet}, but were sensitive to input quality. Extensions to multiple references improved robustness but remained dependent on reliable landmark detection, which often fails under severe degradation and introduces temporal jitter in videos \cite{li2022dmdnet}.

Alternative approaches construct personalized identity spaces in GAN latent domains using numerous subject images \cite{nitzan2022mystyle}, achieving high fidelity at the cost of per user optimization and poor scalability. In the video domain, reference guided solutions are still limited. Existing attempts built on diffusion models \cite{han2025show} often neglect explicit temporal modeling and struggle with data scarcity during fine-tuning. In contrast, our method aligns the degraded video with the reference via geometric cues, enabling robust identity transfer without requiring explicit correspondence. Combined with stochastic conditioning dropout, the framework naturally supports both reference-guided and reference-free inference, achieving a flexible balance between identity fidelity, temporal consistency, and deployment efficiency.

\section{Method}

\begin{figure*}[t]
    \centering
    \includegraphics[width=1.0\linewidth]{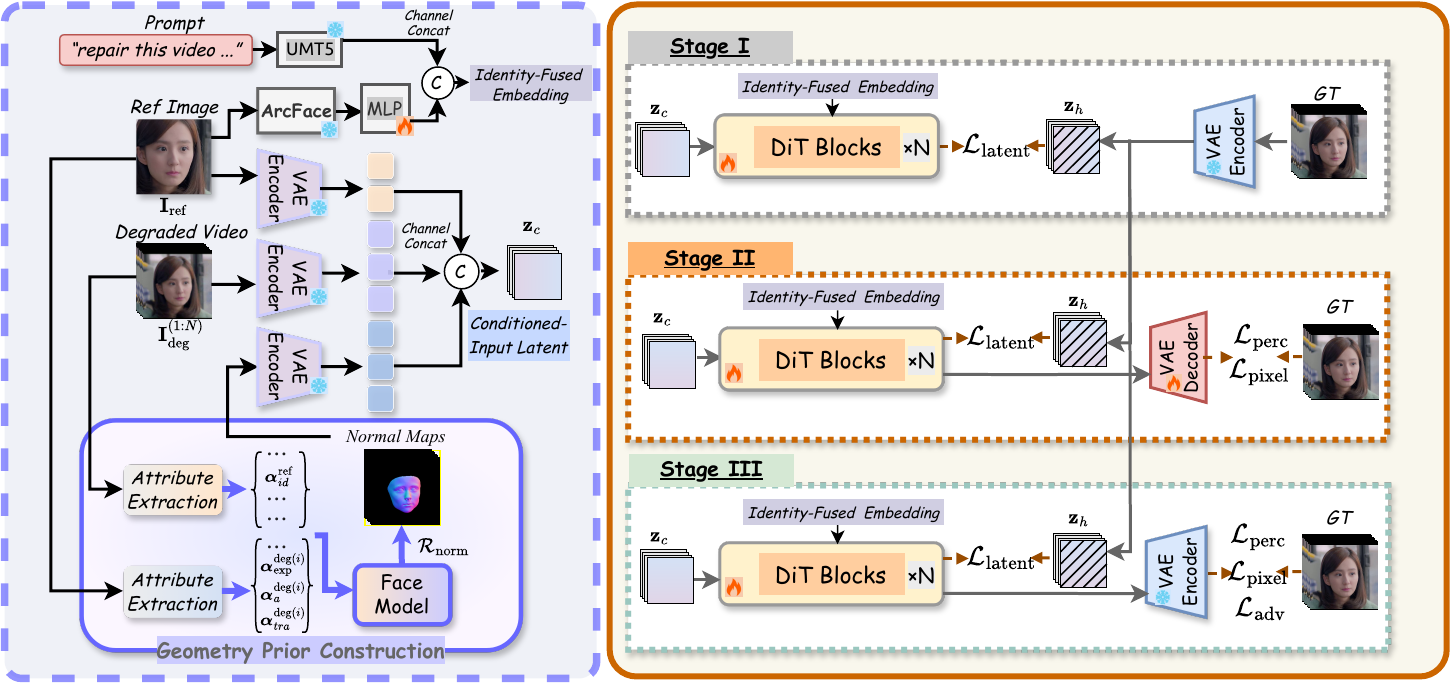}
    \caption{Overview of the proposed \textbf{TIGER} framework, which \textbf{T}ames \textbf{I}dentity, \textbf{G}eometry, and g\textbf{E}nerative p\textbf{R}iors for high-quality FVR. Given a degraded video and a high-quality reference image, we estimate 3D facial parameters to construct geometry-aware constraints that fuse the reference identity with the degraded video’s pose and camera. The resulting geometry prior, together with an identity embedding, conditions a one-step rectified-flow mapper in latent space. 
    We adopt a progressive three-stage optimization strategy: (I) geometry-identity conditioned latent mapping, (II) decoder-coupled detail refinement, and (III) adversarial distribution alignment. }
    \label{fig:pipeline}
\end{figure*}

\subsection{Overview}
As illustrated in Fig.~\ref{fig:pipeline}, we present TIGER, a framework designed to resolve the FVR trilemma by synergizing identity, geometry, and generative priors.
Given a degraded video $\mathbf{I}^{(1:N)}_{\text{deg}}$ and a reference image $\mathbf{I}_{\text{ref}}$, we first extract their facial attributes into a disentangled 3D parameter space. By performing cross-source fusion, we combine the static identity from $\mathbf{I}_{\text{ref}}$ with the dynamic motion from $\mathbf{I}^{(1:N)}_{\text{deg}}$ to render a sequence of geometry-consistent normal maps $\mathbf{I}^{(1:N)}_{\text{norm}}$.
These geometric cues, along with an identity embedding generated by ArcFace~\cite{deng2019arcface}, form the structured priors for our restoration network. We adopt a DiT-based one-step rectified-flow mapper to perform latent transport. By leveraging the generative priors of the pretrained foundation model, TIGER directly maps the degraded latent $\mathbf{z}_l$ to the clean manifold in a single forward pass, ensuring high-fidelity results with maximum efficiency.

\subsection{Geometry Prior Construction}
\label{subsec:geometry_prior}

A major hurdle in FVR is that a static reference image provides identity cues that are fundamentally entangled with its specific viewpoint and expression.
Relying on 2D-based reference guidance often fails when the video subject undergoes significant pose variations, as the fixed-viewpoint prior cannot provide consistent structural constraints across dynamic frames.
To address this, we construct a geometry prior in the disentangled 3D parameter space, where identity- and motion-related factors can be explicitly factorized, enabling consistent identity preservation under dynamic pose variations.

Following the 3D Morphable Model (3DMM)~\cite{3dmm}, the 3D face vertices are parameterized as:
\begin{equation}
    \mathbf{V}_{3d}(\boldsymbol{\alpha}_{id}, \boldsymbol{\alpha}_{\exp}, \boldsymbol{\alpha}_{a}, \boldsymbol{\alpha}_{tra}) = \mathbf{R}(\boldsymbol{\alpha}_a)(\overline{\mathbf{V}} + \boldsymbol{\alpha}_{id}\mathbf{A}_{id} + \boldsymbol{\alpha}_{\exp}\mathbf{A}_{\exp}) + \boldsymbol{\alpha}_{tra},
\end{equation}
where $\overline{\mathbf{V}}$ denotes the mean face shape, $\mathbf{A}_{id}$ and $\mathbf{A}_{exp}$ are the identity and expression bases learned from statistical face models.
The parameters $\boldsymbol{\alpha}_{id}$, $\boldsymbol{\alpha}_{exp}$, $\boldsymbol{\alpha}_a$, and $\boldsymbol{\alpha}_{tra}$ control identity, expression, rotation, and translation, extracted from the pretrained 3DDFA-V3~\cite{wang20243ddfa} respectively. 

\noindent
\textbf{Decoupled Attribute Extraction.}
To obtain the coefficients for identity and motion, we employ a pretrained 3D face reconstructor (3DDFA-V3 \cite{wang20243ddfa}) as a spatial parameter encoder $\mathcal{E}_{3d}$.
For the high-quality reference image $\mathbf{I}_{\text{ref}}$, we extract its static identity representation: $\boldsymbol{\alpha}_{id}^{\text{ref}} = \mathcal{E}_{3d}(\mathbf{I}_{\text{ref}})$.
Simultaneously, for each degraded frame $\mathbf{I}_{\text{deg}}^{(i)}$, we extract the time-varying motion parameters: $\{\boldsymbol{\alpha}_{\exp}^{\text{deg}(i)}, \boldsymbol{\alpha}_{a}^{\text{deg}(i)}, \boldsymbol{\alpha}_{tra}^{\text{deg}(i)}\} = \mathcal{E}_{3d}(\mathbf{I}_{\text{deg}}^{(i)})$.
This step maps raw pixels into a common, disentangled parameter space, enabling the subsequent cross-source fusion.

\noindent
\textbf{Cross-source Parameter Fusion.} 
To construct geometry-aware guidance that preserves identity and tracks motion dynamics, we explicitly decompose facial geometry estimation into complementary identity and motion streams.
This design addresses the limited identity guidance of 2D references: while $\mathbf{I}_{\text{ref}}$ provides clean identity traits, it lacks motion; conversely, $\mathbf{I}^{(1:N)}_{\text{deg}}$ contains the target motion but suffers from identity ambiguity.

Specifically, we decompose the 3DMM parameter estimation into two complementary streams:
\begin{itemize}
    \item \textit{\textbf{Identity from Reference:}} We extract identity coefficients $\boldsymbol{\alpha}_{id}^{\text{ref}}$ from the high-quality $\mathbf{I}_{\text{ref}}$. These coefficients represent view-invariant facial traits and are kept fixed across all frames to ensure rigorous identity preservation.
    \item \textit{\textbf{Motion from Degraded Video:}} For each frame $i \in \{1, \dots, N\}$ in $\mathbf{I}^{(1:N)}_{\text{deg}}$, we estimate the dynamic parameters: pose $\boldsymbol{\alpha}_{a}^{\text{deg}(i)}$, expression $\boldsymbol{\alpha}_{\exp}^{\text{deg}(i)}$, and translation $\boldsymbol{\alpha}_{tra}^{\text{deg}(i)}$. These capture the precise motion dynamics of the input video despite the presence of degradations.
\end{itemize}
By fusing the static identity from the reference with the dynamic motion from the degraded video, we construct a hybrid 3D geometry for each frame:

\begin{equation}
    \mathbf{V}_{3d}^{(i)} = \mathbf{V}_{3d}\big(\boldsymbol{\alpha}_{id}^{\text{ref}}, \boldsymbol{\alpha}_{\exp}^{\text{deg}(i)}, \boldsymbol{\alpha}_{a}^{\text{deg}(i)}, \boldsymbol{\alpha}_{tra}^{\text{deg}(i)}\big).
\end{equation}

Finally, we employ a fixed perspective camera following~\cite{wang20243ddfa} to project the 3D vertices onto the 2D image plane.
We render surface normal maps using the differentiable renderer $\mathcal{R}_{\text{norm}}$, as
\begin{equation}
    \mathbf{I}_{\text{norm}}^{(i)} = \mathcal{R}_{\text{norm}}\big( \mathbf{V}_{3d}^{(i)} \big),
    \label{eq:normal_render}
\end{equation}
These normal maps serve as a view-agnostic geometric prior, providing the restoration network with a structurally consistent scaffold that maintains the subject's identity across diverse poses and expressions.

\subsection{Prior-Conditioned One-Step Rectified Flow}
To bridge the gap between degraded inputs and photorealistic reconstructions, we formulate FVR as a prior-conditioned latent transport problem. Unlike standard diffusion models that rely on stochastic multi-step sampling, we leverage a one-step rectified flow to directly map the degraded distribution to the clean data manifold.

Specifically, given the ground-truth video $\mathbf{I}^{(1:N)}$ and its degraded version $\mathbf{I}^{(1:N)}_{\text{deg}}$, we operate in the latent space of a pretrained video VAE with encoder $\mathcal{E}(\cdot)$ and decoder $\mathcal{D}(\cdot)$. The clean and degraded sequences are encoded as:
\begin{equation}
\mathbf{z}_h = \mathcal{E}(\mathbf{I}^{(1:N)}), \quad \mathbf{z}_l = \mathcal{E}(\mathbf{I}^{(1:N)}_{\text{deg}}),
\end{equation}
where $\mathbf{z}_h \sim p_{\text{data}}$ represents the latent distribution of natural videos. We define the latent residual $\mathbf{y} = \mathbf{z}_h - \mathbf{z}_l$ as the restoration target. 
Following the rectified flow~\cite{rectified_flow}, we construct a linear probability path between the degraded and clean latents:
\begin{equation}
    \textbf{z}_t = (1 - t)\, \textbf{z}_l + t\, \textbf{z}_h, \quad t \in [0,1].
    \label{eq:path}
\end{equation}
The theoretical velocity of this path is constant:
$\textbf{v}_t=\frac{d \textbf{z}_t}{d t}=\textbf{z}_h-\textbf{z}_l$.
The velocity predictor $\mathbf{v}_\theta$ is implemented using a Diffusion Transformer (DiT).

To provide dense geometric and visual constraints frame-by-frame, we encode geometry-aware normal maps $\mathbf{I}^{(1:N)}_{\text{norm}}$ and the reference image $\mathbf{I}_{\text{ref}}$ into the latent space. These are concatenated with the degraded latent $\mathbf{z}_l$ along the channel dimension, forming a spatially-aligned input 
\begin{equation}
\mathbf{z}_c = [\mathbf{z}_l, \mathcal{E}(\mathbf{I}^{(1:N)}_{\text{norm}}), \mathcal{E}(\mathbf{I}_{\text{ref}})].
\end{equation}
This ensures that the velocity $\mathbf{v}_\theta$ is strictly bounded by the 3D geometry and reference appearance.
To further anchor the subject's identity, we leverage the projected ArcFace~\cite{deng2019arcface} embedding $\mathbf{f}_{\text{arc}}$ extracted from $\textbf{I}_{\text{ref}}$. We fuse the identity embedding with text embeddings to construct the identity-fused embedding $\mathbf{f}_{\text{id}}$, which is injected into the DiT blocks via the cross-attention mechanism. This design forces the generative process to attend to identity features for eliminating identity drift.

Instead of relying on unguided velocity estimation, we incorporate the proposed priors to rectify the transport direction as:
\begin{equation}
\hat{\mathbf{z}}_h = \mathbf{z}_l + \mathbf{v}_{\theta} \bigl( \mathbf{z}_c, \mathbf{f}_{\text{id}}, t^* \bigr),
\label{eq:onestep_refined}
\end{equation}
where $t^*$ is the fixed rectification time (set to $0.4$ for both training and inference).

\subsection{Progressive Three-Stage Optimization}

To synergize the structured priors and the generative capabilities of the foundation model, we adopt a coarse-to-fine optimization strategy that progressively refines identity fidelity, textural detail, and holistic realism.

\noindent
\textbf{Stage I: Geometry-Identity Conditioned Latent Mapping.}
The initial stage focuses on establishing a robust one-step mapping from the degraded latent $\mathbf{z}_l$ to the clean manifold $\mathbf{z}_h$. We feed the spatially-aligned input $\mathbf{z}_c$ into the DiT backbone, while anchoring the identity via cross-attention with $\mathbf{f}_{\text{arc}}$. The model is optimized by minimizing the latent-space MSE:
\begin{equation}
    \mathcal{L}_{\text{latent}} = \mathbb{E}\!\left[\left\| \hat{\textbf{z}}_h - \textbf{z}_h \right\|_2^2\right].
\end{equation}
The training objective of Stage I is defined as $\mathcal{L}_{\text{Stage1}}=\mathcal{L}_{\text{latent}}$. This stage tames the model to reconstruct basic facial structures and identity consistent with the guidance.

\noindent\textbf{Stage II: Decoder-Coupled Detail Refinement.}
While Stage I achieves structural restoration, the pretrained VAE decoder $\mathcal{D}$ may introduce texture softening when processing restored latents. 
To alleviate this, we decode \(\hat{\textbf{z}}_h\) into frames \(\hat{\textbf{I}}^{(1:N)}_{\text{rec}} = \mathcal{D}(\hat{\textbf{z}}_h)\) and jointly optimize DiT and $\mathcal{D}$ with a perceptual loss (LPIPS~\cite{zhang2018lpips}) in addition to latent and pixel losses:
\begin{equation}
    \mathcal{L}_{\text{Stage2}}=\lambda_1 \mathcal{L}_{\text{latent}}+\lambda_2 \mathcal{L}_{\text{perc}}+\lambda_3 \mathcal{L}_{\text{pixel}}.
\end{equation}
Where $\mathcal{L}_{pixel}$ is the MSE loss between the output and the ground truth. This strengthens texture and edge fidelity while preserving identity and pose accuracy.

\noindent\textbf{Stage III: Adversarial Distribution Alignment.}
To bridge the perceptual gap, we employ adversarial training with a curated face-video dataset to better fit the distribution of real portrait videos. A spatiotemporal discriminator \(\mathcal{D}_{\phi}\) is employed to supervise the realism and temporal coherence. The overall objective is:
\begin{equation}
    \mathcal{L}_{\text{Stage3}}=\lambda_1 \mathcal{L}_{\text{latent}}+\lambda_2 \mathcal{L}_{\text{perc}}+\lambda_3 \mathcal{L}_{\text{pixel}}+\lambda_4\, \mathcal{L}_{\text{adv}}.
\end{equation}
This improves global coherence and perceptual quality while maintaining identity faithfulness and temporal stability.

To improve robustness, we apply conditional dropout to \(\textbf{z}_{\text{c}}\) during training: with a certain probability, we drop the identity embedding or the geometry control stream, making the model resilient to missing or unreliable priors at inference. At the inference time, the model can operate with reference guidance when available, or fall back to geometry-only or unguided restoration under challenging conditions.

\subsection{Quality-Aware Face-Centric Dataset Construction}

\begin{figure}[t]
    \centering
    \includegraphics[width=\linewidth]{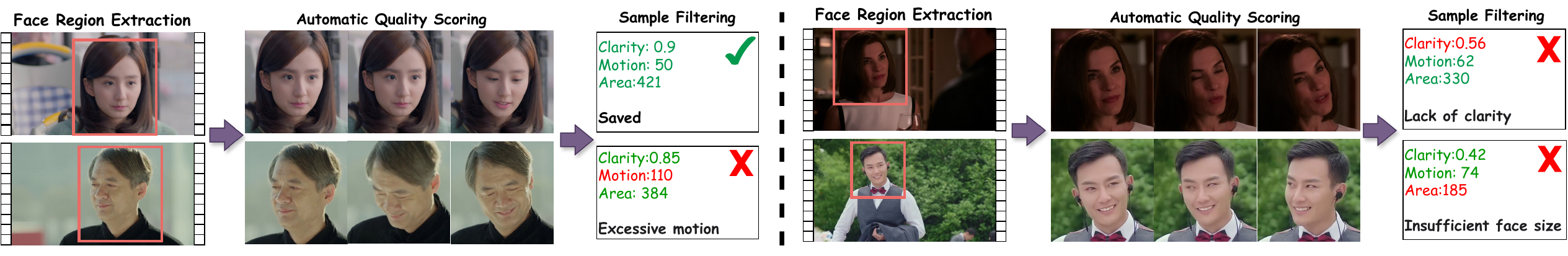}
    \caption{Pipeline of the proposed high-quality face video filtering with three core dimensions: 
    (1) motion amplitude, (2) face clarity, and (3) face proportion.}
    \label{fig:data_pipeline}
\end{figure}

\noindent
Although existing mainstream face video datasets such as VFHQ~\cite{xie2022vfhq} and CelebVHQ~\cite{zhu2022celebvhq} contain high-resolution portrait videos, they still exhibit notable limitations for FVR. 
First, the facial regions often occupy a limited proportion of each frame, causing models to implicitly emphasize background reconstruction at the expense of fine-grained facial textures. 
Second, many samples are affected by motion blur, low illumination, or compression artifacts. 
Without reliable high-frequency facial details as supervision, models are prone to generating spurious textures during inference, thereby degrading restoration fidelity.

To address these challenges, we construct a quality-aware, face-centric training dataset via a dedicated video selection pipeline (Fig.~\ref{fig:data_pipeline}). We collect raw videos from three public datasets: OpenHumanVid \cite{li2025openhumanvid}, CelebV-HQ~\cite{zhu2022celebvhq}, and VFHQ~\cite{xie2022vfhq}. 
In the quality filtering stage, we establish quantitative criteria across three core dimensions: motion amplitude, face clarity, and face proportion. 
Motion amplitude is quantified by facial centroid displacement, with a threshold of less than 100 pixels to ensure temporal stability. 
Face clarity is evaluated via normalized Laplacian variance, scaled to $[0,1]$ using a logarithmic function with a variance upper bound of $1000$. Samples are filtered to retain only those with a clarity score above $0.7$, thereby excluding blurry frames.
Face proportion requires the minimum side length of the facial region to exceed 320 pixels, ensuring that faces occupy a significant proportion of frames and provide sufficient texture information. Through this process, we provide high-quality training data with geometric consistency, rich details, and controllable degradation levels for FVR.

After filtering, we retain 28,491 high-quality face video samples for training. Reference images are obtained by random sampling from video frames followed by background removal, ensuring that the model focuses on learning geometric priors and reconstructing facial details.
The resulting dataset provides high-resolution, geometrically consistent, and detail-rich supervision tailored for FVR.

\section{Experiments}

\subsection{Implementation details}

\textbf{Training Details.} We utilize Wan2.1-Fun-1.3B \cite{wan2025wan} as the base model. The DiT backbone is fine-tuned via LoRA \cite{hu2022lora} with a rank of 384, whereas the VAE decoder is optimized with full-parameter fine-tuning. Training is conducted for 10 epochs on 8 GPUs with a batch size of 8. For all training phases, we set the learning rate to $10^{-4}$ and adopt the Adam \cite{adam2014method} optimizer. In Stage I, each video clip consists of 81 frames. 
In Stages II and III, the clip length is set to 21 frames to reduce computational overhead. 
All videos are resized to a spatial resolution of $512\times512$ for training. 

\noindent
\textbf{Datasets.} 
For Stage I, we use 28,491 filtered facial video samples from three datasets (OpenHumanVid \cite{li2025openhumanvid}, CelebV-HQ \cite{zhu2022celebvhq}, and VFHQ\cite{xie2022vfhq}) to train the model’s ability to adhere to geometry priors. The reference images are sampled randomly from video frames with background removed. Following common practices \cite{xie2025bfvr}, Stages II and III are trained on filtered video samples from VFHQ. In addition to the standard 50 test samples from VFHQ-Test, we construct a new test benchmark for reference-guided FVR without ground truth (GT) based on the VoxCeleb2 \cite{chung2018voxceleb2} dataset. This benchmark consists of 100 low-quality real-world video samples and corresponding extra-scene photos of the same subjects as reference images. 

\noindent
\textbf{Evaluation Metrics.} 
For synthetic VFHQ-Test dataset with GT, we evaluate performance using PSNR, SSIM, LPIPS \cite{zhang2018lpips}, and FVD \cite{unterthiner2019fvd}. For real-world datasets, we adopt CLIP-IQA \cite{wang2023clipiqa}, MUSIQ \cite{ke2021musiq}, and LIQE \cite{zhang2023liqe}. 
Furthermore, we use IDS (identity similarity distance) to assess identity preservation, which is defined as the cosine similarity with ArcFace \cite{deng2019arcface}. 
To measure inter-frame identity consistency, we propose VIRD (video identity and reference distance), representing the identity similarity between each frame in the facial video and the reference image. The metric is defined as:
\begin{equation}
\small
    \text{VIRD} = \frac{1}{N} \sum_{i=1}^{N} \|\mathcal{F}_{Arc}(I_i), \mathcal{F}_{Arc}(I_{ref})\|_2,
    \label{eq:vird}
\end{equation}
where we use ArcFace \cite{deng2019arcface} as the feature extractor $\mathcal{F}_{Arc}(\cdot)$ to obtain discriminative identity embeddings.

\definecolor{best}{RGB}{255,230,230}
\definecolor{second}{RGB}{230,240,255}

\begin{table}[t!]
\centering
\setlength{\tabcolsep}{1.1mm}{
\caption{Quantitative comparison results on the VFHQ-Test dataset. 
$\uparrow$ / $\downarrow$ denote higher / lower is better, respectively.
We highlight the \colorbox{best}{best} and \colorbox{second}{second} metrics. 
}
\label{tab:quantitative_comparison}
\scalebox{0.76}{
\begin{tabular}{l c c c c c c c c c c}
\toprule
Method & {PSNR $\uparrow$} & {SSIM $\uparrow$} & {LPIPS $\downarrow$} & {IDS $\uparrow$} & {FVD $\downarrow$} & {LIQE $\uparrow$} & {MUSIQ $\uparrow$} & {CLIP-IQA $\uparrow$} & {VIRD $\downarrow$} \\
\midrule

DOVE & \colorbox{second}{30.4082} & \colorbox{second}{0.8751} & \colorbox{second}{0.1189} & \colorbox{second}{0.9283} & \colorbox{second}{103.892} & 3.7983 & \colorbox{second}{69.9454} & 0.4654 & \colorbox{second}{0.4891} \\
BasicVSR++  & 27.9281 & 0.8216 & 0.2822 & 0.8970 & 325.657 & 1.6914 & 43.9436 & 0.3498 & 0.5355 \\
STAR  & 24.6980 & 0.7821 & 0.2307 & 0.8259 & 251.939 & \colorbox{second}{3.8819} & 68.4678 & \colorbox{best}{0.5889} & 0.6481 \\
SeedVR2-3B  & 23.1543 & 0.7848 & 0.1638 & 0.8931 & 593.618 & 3.7019 & 67.8074 & \colorbox{second}{0.5076} & 0.5438 \\

\midrule
KEEP & 28.0798 & 0.8488 & 0.1776 & 0.7682 & 370.081 & 2.9652 & 61.3693 & 0.4097 & 0.7430 \\
BFVR & 26.8917 & 0.8377 & 0.2128 & 0.8454 & 316.215 & 1.8835 & 53.3688 & 0.3725 & 0.6265 \\
PGTFormer & 28.7534 & 0.8466 & 0.2052 & 0.8421 & 334.736 & 2.9345 & 64.5613 & 0.5059 & 0.6310 \\
SVFR & 25.6862 & 0.8019 & 0.1936 & 0.8943 & 147.999 & 3.2908 & 65.8764 & 0.4588 & 0.5496 \\

\midrule
\textbf{TIGER(Ours)} & \colorbox{best}{30.7289} & \colorbox{best}{0.8773} & \colorbox{best}{0.0697} & \colorbox{best}{0.9481} & \colorbox{best}{40.4222} & \colorbox{best}{4.0919} & \colorbox{best}{70.1086} & 0.4756 & \colorbox{best}{0.4536} \\

\bottomrule
\end{tabular}}}
\end{table}

\subsection{Comparison With State-of-the-Art}

\textbf{Compared Methods.} We conducted a comparison with current state-of-the-art video enhancement methods, including BasicVSR++\cite{chan2022basicvsr++}, DOVE\cite{chen2025dove}, STAR\cite{Xie_2025_STAR}, and SeedVR2-3B\cite{wang2025seedvr2}, as well as blind FVR algorithms KEEP\cite{feng2024keep}, BFVR\cite{xie2025bfvr}, PGTFormer\cite{pgtformer}, and SVFR\cite{wang2025svfr}. Among these, DOVE, STAR, SeedVR2-3B, and SVFR are all built upon generative priors of video generation models, leveraging their capacity to model complex video distributions and recover high-quality facial details from degraded inputs.

\noindent
\textbf{Quantitative Results.} Our method demonstrates superior performance over SoTAs across both VFHQ-Test and VoxCeleb2 datasets, excelling in critical metrics that measure visual quality, perceptual fidelity, identity preservation, and temporal consistency. On VFHQ-Test, it outperforms SoTAs in eight of nine metrics—significantly improving on LPIPS, IDS, and temporal stability (FVD, VIRD)—while establishing new benchmarks in visual quality (PSNR, SSIM) and LIQE. Similarly, on VoxCeleb2, it dominates three metrics, showcasing exceptional perceptual quality (LIQE, MUSIQ) and consistency (VIRD). Collectively, these results validate its efficacy in balancing identity retention, temporal coherence, and visual realism, establishing its superiority across FVR task.

\begin{figure}[t]
    \centering
    \includegraphics[width=\linewidth]{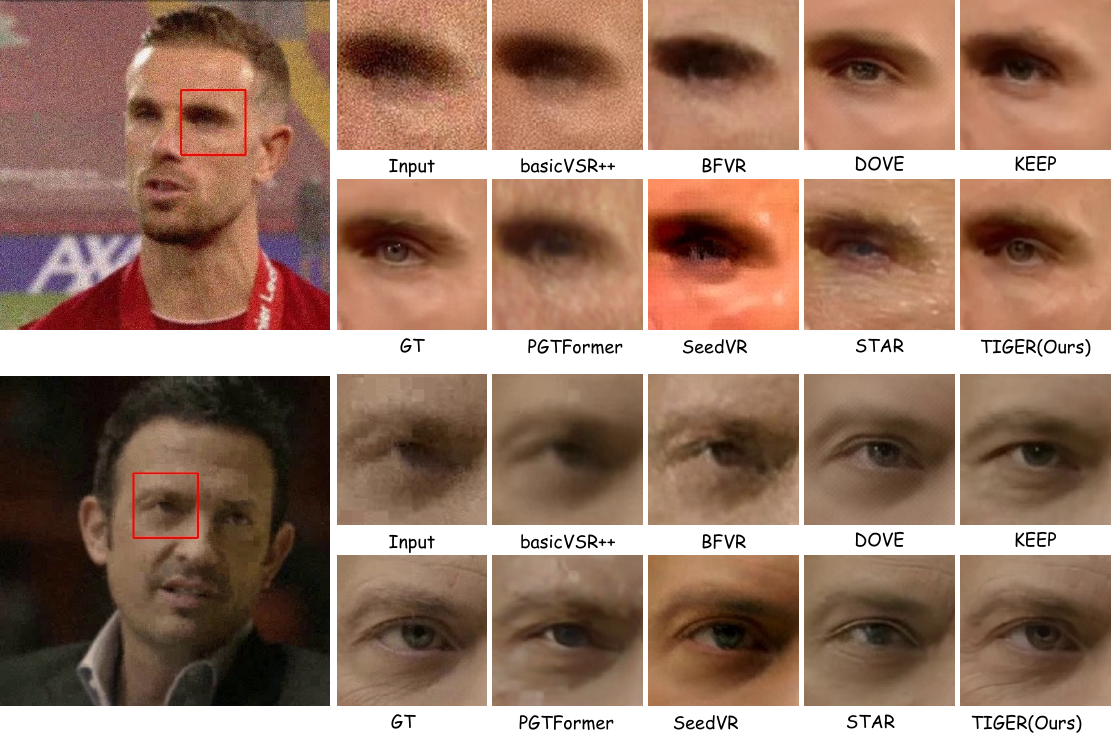}
    \caption{Qualitative comparison with other state-of-the-art models. Our approach better preserves identity-discriminative features and restores more fine-grained facial details.}
    \label{fig:qualitative}
\end{figure}

\begin{figure}[t]
\centering
\begin{minipage}{0.42\linewidth}
    \centering
    \captionof{table}{Quantitative comparison of our method with SoTA approaches on the VoxCeleb2 dataset. 
    }
    \label{tab:vox2_comparison}
    \scalebox{0.65}{
    \begin{tabular}{l c c c c}
    \toprule
    Method & {LIQE $\uparrow$} & {MUSIQ $\uparrow$} & {CLIP-IQA $\uparrow$} & {VIRD $\downarrow$} \\
    \midrule
    DOVE & 2.0396 & 58.8717 & 0.3731 & \colorbox{second}{0.9360} \\
    BasicVSR++ & 1.0208 & 37.3428 & 0.2847 & 0.9470 \\
    STAR & \colorbox{second}{2.8615} & \colorbox{second}{59.9601} & \colorbox{best}{0.5706} & 0.9625 \\
    SeedVR & 1.3459 & 48.9973 & 0.3143 & 0.9408 \\
    \midrule
    KEEP & 2.1701 & 55.3556 & 0.3630 & 0.9767 \\
    BFVR & 1.0706 & 43.0584 & 0.3191 & 0.9819 \\
    PGTFormer & 1.8864 & 56.8901 & 0.4370 & 0.9641 \\
    SVFR & 2.1780 & 57.7379 & 0.4031 & 0.9536 \\
    \midrule
    \textbf{TIGER(Ours)} & \colorbox{best}{3.7412} & \colorbox{best}{66.2382} & \colorbox{second}{0.4933} & \colorbox{best}{0.9322} \\
    \bottomrule
    \end{tabular}}
\end{minipage}
\hfill
\begin{minipage}{0.55\linewidth}
    \centering
    \includegraphics[width=\linewidth]{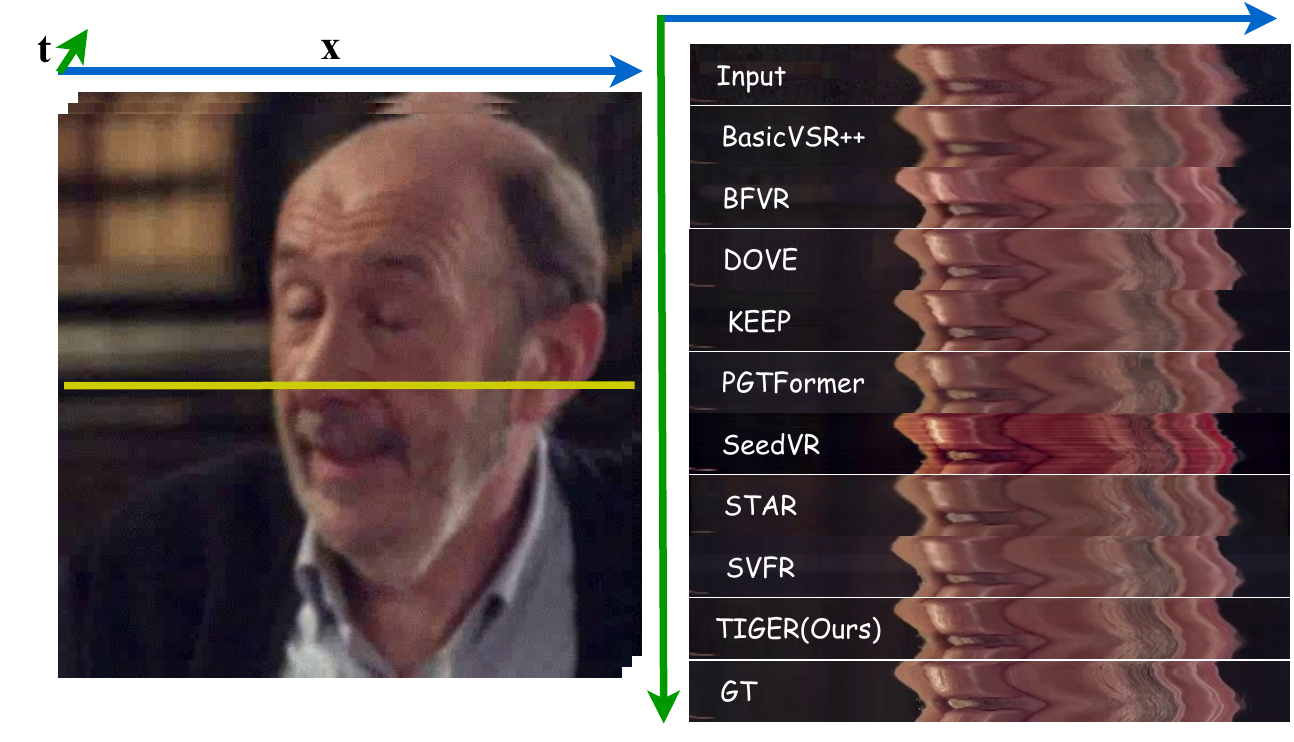}
    \captionof{figure}{Qualitative comparison of temporal consistency, stacking the yellow line across frames.}
    \label{fig:qualitative_time}
\end{minipage}
\end{figure}

\noindent
\textbf{Qualitative Results.} 
We provide visual comparisons in Fig. \ref{fig:qualitative}. 
Our method produces more realistic and coherent results compared to other approaches. For instance, our model successfully reconstructs fine details such as skin texture and facial features, while other methods often produce overly smooth or distorted results. Similarly, in complex motion scenarios, our approach maintains sharper restoration with better preservation of dynamic elements across frames.

We also visualize the temporal profile in Fig. \ref{fig:qualitative_time}, demonstrating that our method achieves excellent temporal consistency even under challenging conditions. Existing approaches often struggle with complex degradations, exhibiting visible artifacts such as misalignment or blurring between consecutive frames. This results in a more natural and visually pleasing restoration that better aligns with human perception.

\subsection{Ablation Studies}

We conduct ablation studies to validate the effectiveness of our approach. Specifically, we evaluate the impact of reference priors, the contribution of each training stage, cross-source parameter fusion, and identity conditioning on VFHQ-Test. Additionally, the loss weight balancing strategy is assessed on VoxCeleb2.

\noindent
\textbf{Effectiveness of Different Priors.} 
We analyze the contributions of the identity and geometry priors. 
``w/o Ide.'' removes the identity input, while ``w/o Geo.'' disables the 3D normal-based geometry prior. 
VIRD evaluates identity fidelity, LPIPS measures perceptual similarity, and FVD reflects overall video consistency. As shown in Tab.~\ref{tab:abl_time}, removing both priors leads to the worst performance, with degraded identity preservation and visual quality. Using only a reference image improves results significantly, as it provides strong identity cues through identity embeddings. However, without explicit geometric guidance, the model lacks structural consistency across poses and expressions. The best performance is achieved when both reference and geometry priors are employed, demonstrating their complementary roles: the reference anchors identity, while geometry enforces cross-frame structural coherence. 

\begin{figure}[t]
\centering
\begin{minipage}{0.49\linewidth}
    \centering
    \setlength{\tabcolsep}{2mm}{
    \captionof{table}{Ablation study on the effectiveness of identity (Ide.) and geometry (Geo.) priors.
    }
    \label{tab:abl_time}
    \scalebox{0.8}{
    \begin{tabular}{l c |c c c}
    \toprule
    Ide. & Geo. & {VIRD $\downarrow$} & {FVD $\downarrow$} & {LPIPS $\downarrow$} \\
    \midrule
    $\times$ & $\times$  & 0.4940 & 101.5778 & 0.1942 \\
    $\times$ & $\checkmark$ & 0.4791 & 70.4368 & 0.1275 \\
    $\checkmark$ & $\times$ & 0.4783 & 71.1703 & 0.1272 \\
    $\checkmark$ & $\checkmark$ & \textbf{0.4492} & \textbf{40.4222} & \textbf{0.0697} \\
    \bottomrule
    \end{tabular}}}
\end{minipage}
\hfill
\begin{minipage}{0.49\linewidth}
    \centering
    \setlength{\tabcolsep}{2mm}{
    \captionof{table}{Ablation study on the effectiveness of the progressive three-stage optimization. 
    }
    \label{tab:abl_stage}
    \scalebox{0.8}{
    \begin{tabular}{ccc|c c c}
    \toprule
    I & II & III & {VIRD $\downarrow$} & {FVD $\downarrow$} & {LPIPS $\downarrow$} \\
    \midrule
    $\checkmark$ & & & 0.4774 & 70.0517 & 0.1261 \\
    $\checkmark$ & $\checkmark$ &   & 0.4628 & 60.5291 & 0.0911 \\
    $\checkmark$ & $\checkmark$ & $\checkmark$ & \textbf{0.4492} & \textbf{40.4222} & \textbf{0.0697} \\
    \bottomrule
    \end{tabular}}}
\end{minipage}
\end{figure}


\noindent
\textbf{Effectiveness of Three Stage Pipeline.}  
We conduct a ablation study by incrementally introducing each stage of the proposed three-stage training pipeline. Specifically, we evaluate: (i) training with only Stage I, (ii) Stage I + Stage II, and (iii) the full three-stage pipeline. As shown in Tab.~\ref{tab:abl_stage}, Stage I training only the latent mapper yields limited perceptual quality despite reasonable identity preservation. In Stage II, joint fine-tuning of the mapper and the VAE decoder significantly improves visual fidelity by refining textures and edges. To stabilize adversarial training in Stage III, we freeze the decoder parameters to prevent mode collapse and performance degradation, while optimizing the mapper with GAN loss. This final stage further boosts both identity consistency and perceptual quality, demonstrating the necessity of staged optimization.

\begin{figure}[t]
    \centering
    \includegraphics[width=0.96\linewidth]{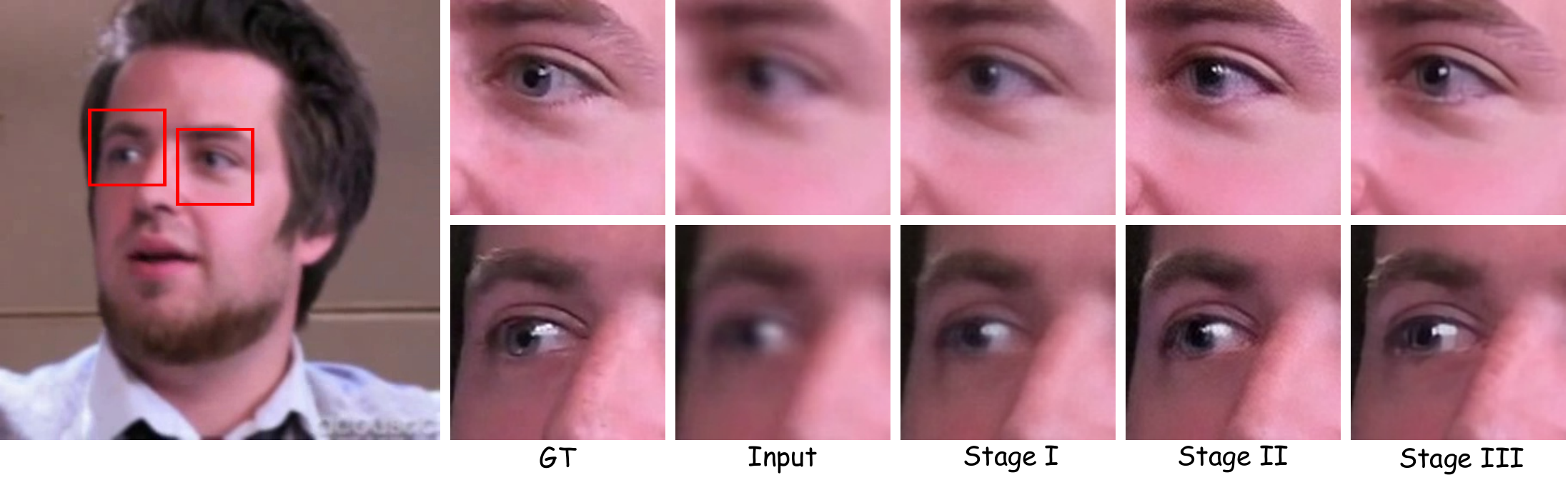}
    \caption{Qualitative of ablation study on the three-stage training pipeline.}
    \label{fig:abl_stage}
\end{figure}

\noindent
\textbf{Impact of Cross-source Parameter Fusion.} 
We ablate the parameter sources for geometry construction in Tab.~\ref{tab:abl_cross-source}.
Using parameters solely from the degraded video leads to unreliable identity cues due to the severe degradation.
Conversely, extracting both from the reference images provides stable identity but fails to capture the true motion dynamics, resulting in degraded performance.
Disentangling identity and motion in the 3D parameter space and recombining them from complementary sources provides more reliable geometric guidance for FVR.

\noindent
\textbf{Impact of Identity Conditioning.}
We evaluate the impact of introducing identity embeddings into the cross-attention layers.
As shown in Tab.~\ref{tab:abl_cross-ide}, compared to the text-only baseline, incorporating identity embeddings significantly improves overall performance.

\begin{figure}[t]
\centering
\begin{minipage}{0.48\linewidth}
    \centering
    \setlength{\tabcolsep}{2mm}{
    \captionof{table}{Effectiveness of cross-source parameter fusion. \textit{ref} and \textit{deg} indicate parameters extracted from reference image and degraded video, respectively.}
    \label{tab:abl_cross-source}
    \scalebox{0.7}{
    \begin{tabular}{cc |c c c}
    \toprule
    \makecell{Identity \\ Source} & \makecell{Motion \\ Source} & {VIRD $\downarrow$} & {FVD $\downarrow$} & {LPIPS $\downarrow$} \\
    \midrule
    \textit{deg} & \textit{deg} & 0.4790 & 71.0134 & 0.1275 \\
    \textit{ref} & \textit{ref} & 0.4781 & 71.3495 & 0.1274 \\
    \textit{ref} & \textit{deg} & \textbf{0.4492} & \textbf{40.4222} & \textbf{0.0697} \\
    \bottomrule
    \end{tabular}}}
\end{minipage}
\hfill
\begin{minipage}{0.48\linewidth}
    \centering
    \setlength{\tabcolsep}{1.4mm}{
    \captionof{table}{Ablation on the identity-fused embedding.}
    \label{tab:abl_cross-ide}
    \scalebox{0.7}{
    \begin{tabular}{cc |c c c}
    \toprule
    \makecell{Text \\ Embedding} & \makecell{Identity \\ Embedding} & {VIRD $\downarrow$} & {FVD $\downarrow$} & {LPIPS $\downarrow$} \\
    \midrule
    $\times$ & $\checkmark$ & 0.6122 & 618.2963 & 0.3666 \\
    $\checkmark$ & $\times$ & 0.4501 & 41.0479 & 0.0698 \\
    $\checkmark$ & $\checkmark$ & \textbf{0.4492} & \textbf{40.4222} & \textbf{0.0697} \\
    \bottomrule
    \end{tabular}}}
\end{minipage}
\end{figure}

\noindent
\textbf{Impact of Loss Weights.}
Tab.~\ref{tab:alb_weight} presents the ablation results under different weight configurations. 
Stronger latent supervision or weaker pixel loss degrades identity consistency, while increasing the adversarial weight improves perceptual quality at the cost of slight fidelity drop. 
The configuration $(1.0, 2.0, 10.0, 0.2)$ achieves the best trade-off between identity preservation and visual realism and is thus adopted.

\begin{table}[t!]
\centering
\setlength{\tabcolsep}{2mm}  
\captionof{table}{Impact of loss weight configurations ($\lambda_1$, $\lambda_2$, $\lambda_3$, $\lambda_4$) on latent supervision, perceptual similarity, pixel reconstruction, and adversarial learning.}
\label{tab:alb_weight}
\scalebox{0.8}{
\begin{tabular}{cccc|cc}
\toprule
$\lambda_1$ & $\lambda_2$ & $\lambda_3$ & $\lambda_4$ & VIRD $\downarrow$ & LIQE $\uparrow$ \\
\midrule
1.0 & 2.0 & 10.0 & 0.1 & \textbf{0.9280} & 3.0988 \\
2.0 & 2.0 & 10.0 & 0.1 & 0.9315 & 2.8255 \\
1.0 & 2.0 & 5.0 & 0.1 & 0.9356 & 3.1156 \\
1.0 & 3.0 & 10.0 & 0.1 & 0.9452 & 3.1068 \\
1.0 & 2.0 & 10.0 & 0.2 & 0.9322 & \textbf{3.7412} \\
\bottomrule
\end{tabular}
}
\end{table}

\section{Conclusion}

In this work, we introduced TIGER, a diffusion-based, one-step inference framework for FVR that excels in identity preservation, temporal consistency, and visual quality. Our approach synergistically \textbf{T}aming \textbf{I}dentity, \textbf{G}eometry, and g\textbf{E}nerative p\textbf{R}iors to produce high-fidelity, coherent results. To mitigate identity drift, we anchor the restoration process using robust identity embeddings from a reference image. Coherence across frames is enforced by incorporating 3D geometric priors as structural constraints. The one-step inference design, augmented with adversarial training for texture refinement, ensures both speed and photorealism. Furthermore, we propose a three-stage training strategy and a new large-scale benchmark dataset to facilitate stable optimization and rigorous evaluation. Extensive experiments confirm that TIGER sets a new state-of-the-art in reference-guided FVR. Future work will focus on expanding our training data to encompass a wider range of real-world variations, pushing the boundaries of photorealistic FVR.

\bibliography{ref}


\end{document}